\documentclass{article}
\usepackage{graphicx}
\usepackage[utf8]{inputenc}
\usepackage{amsmath}
\usepackage{amsfonts}
\usepackage{cite}
\usepackage{url}
\usepackage{diagbox}
\usepackage{comment}
\usepackage{xcolor}
\usepackage{authblk}
\usepackage{float}

\title{Hybrid Architecture for Real-Time Video Anomaly Detection: Integrating Spatial and Temporal Analysis}
\author{Fabien Poirier}
\affil{Paris 8 University, LIASD, France}

\begin{document}

\maketitle

\section{Abstract}

In this paper, we propose a new architecture for real-time anomaly detection in video data, inspired by human behavior combining spatial and temporal analyses. This approach uses two distinct models: (i) for temporal analysis, a recurrent convolutional network (CNN + RNN) is employed, associating VGG19 and a GRU to process video sequences; (ii) regarding spatial analysis, it is performed using YOLOv7 to analyze individual images. 
These two analyses can be carried out either in parallel, with a final prediction that combines the results of both analysis, or in series, where the spatial analysis enriches the data before the temporal analysis. Some experimentations are been made to compare these two architectural configurations with each other, and  evaluate the effectiveness of our hybrid approach in video anomaly detection. \vspace{0.5\baselineskip}

\noindent\textbf{Keywords:} Anomaly Detection, Real-Time, YOLOv7, VGG-GRU, Video Analysis, Modular Architecture

\section{Introduction}
\label{sec:introduction}
Anomaly detection in videos is essential for various applications, ranging from security surveillance to disaster management, and the monitoring of large-scale events such as the Olympic Games. In such situations, the ability to quickly identify anomalies can have significant consequences, whether it is ensuring the safety of people or responding effectively to critical situations. 
However, traditional detection systems often rely on an isolated analysis of the temporal aspects of videos, limiting their ability to detect anomalies efficiently in complex environments (cf. Section~\ref{sec:relatedWork}). Since videos are multimodal, incorporating both static visual information (images) and dynamic information (sequences of images), it is necessary to adopt an approach capable of capturing the full richness of potential anomalies as efficiently and quickly as possible. To this end, we propose a hybrid architecture that combines spatial analysis, to detect objects and visual patterns in each image, with temporal analysis to model the dynamics of video sequences. 
For this purpose, we use YOLOv7~\cite{wang2022yolov7}, a state-of-the-art object detection model, coupled with a combination of the supervised learning model VGG19 and GRU to model temporal sequences. This approach enables not only the detection of anomalies based on the presence of suspicious objects but also the identification of suspicious behaviors over time. The originality of our approach lies in the integration of these two types of analysis. We explore two main configurations: a parallel approach (where spatial and temporal analyses are performed simultaneously and combined for a final prediction) against a sequential approach (where spatial analysis enriches the temporal analysis). This integration improves the precision and reliability of anomaly detection by leveraging the strengths of each type of analysis. This paper aims to evaluate the effectiveness of these hybrid configurations by comparing the parallel and sequential approaches, while also analyzing the impact of spatial analysis integration into our architecture. We trained and tested both approaches on a proprietary dataset. The results allow us to assess the specific contribution of spatial analysis, highlighting its strengths and limitations within the context of anomaly detection.

\section{Related works}
\label{sec:relatedWork}
Anomaly detection in videos has made significant advances in recent years~\cite{samaila2024}, driven by progress in deep learning technologies and data analysis techniques. 
In this study, Samaila et al.~\cite{samaila2024} highlights that deep learning has now emerged as a dominant approach compared to traditional machine learning methods. Among the various learning strategies used for video anomaly detection, reinforcement learning remains significantly less explored than supervised and unsupervised learning. 
Consequently, the authors focus primarily on deep learning techniques and these two prevalent learning paradigms. 
Therefore, this section examines recent approaches in object and anomaly detection, as well as hybrid models that combine spatial and temporal analyses.

\subsection{Object Detection}
\label{subsec:objectDetection}
The evolution of object detection models has played a crucial role in improving anomaly detection systems. Zou et al. (2019) in~\cite{zou2019object} published a detailed review of the evolution of object detection techniques over the past two decades. 
Initial methods, such as the Recurrent Convolutional Neural Network (RCNN)~\cite{girshick2014rich}, paved the way for modern object detection but suffered from significant limitations. 
RCNN required generating a large number of region proposals, which were then processed individually by a convolutional network, resulting in extremely high computational costs and slow inference times. 
To address these inefficiencies, Fast RCNN~\cite{girshick2015fast} introduced a more integrated approach, while Faster RCNN~\cite{ren2015faster} replaced region proposal mechanisms with a Region Proposal Network (RPN), significantly accelerating the process. 
Despite these advances, the RCNN family still struggled to meet the demands of real-time detection, especially in scenarios requiring high-speed analysis. This limitation led to the development of the You Only Look Once (YOLO) series~\cite{redmon2016you, redmon2017yolo9000, redmon2018yolov3}, which redefined object detection by formulating it as a single regression problem. YOLO processes an image in a single pass, dividing it into a grid and simultaneously predicting bounding boxes and class probabilities, achieving unparalleled speed and efficiency. 
Successive versions of YOLO have brought significant improvements: YOLOv4 introduced advanced data augmentation techniques~\cite{bochkovskiy2020yolov4}. YOLOv7, currently the latest version available at the time of this publication, incorporates the YOLOR (You Only Learn One Representation) architecture~\cite{wang2021you} and eliminates anchor boxes, enabling ultra-fast image analysis~\cite{wang2022yolov7}.

\subsection{Anomaly Detection}
\label{subsec:anomlyDetection}
Anomaly detection technologies have also evolved and diversified. As for methods focusing solely on temporal analysis, we find technologies such as LSTMs, GRUs, CNNs, and GANs, which have been widely used for anomaly detection in videos, as mentioned by Samaila et al.~\cite{samaila2024}. 
Among these techniques, less common models like C3D (3D Convolutional Networks) have also shown promising results for extracting spatio-temporal features. This technology was previously used by Du Tran et al.~\cite{tran2015learning} for action recognition, addressing problems similar to those discussed here. 
Lin Wang et al.~\cite{wang2023attention} adopted this architecture to propose a weakly-supervised anomaly detection method, using a multi-instance pseudo-label generator and an anomaly detector enhanced by attention. Their goal is to overcome the limitations of traditional anomaly detection methods, particularly the lack of labeled data, through a weakly-supervised approach. The pseudo-label generator produces approximate labels for anomalous videos, transforming anomaly detection into a supervised learning problem. Videos are first processed to extract spatio-temporal features using the C3D network, which serves as a feature encoder. 
The model also integrates attention modules to focus on the anomalous regions of the videos. Finally, it is trained using the generated pseudo-labels and normal videos, and the C3D network parameters are fine-tuned to adapt to the task-specific features. 
Futhermore, Vision Transformers (ViT), although more recent, are being explored for their ability to capture complex relationships in video sequences. 
In their paper, Waseem Ullah et al.~\cite{ullah2023vision} present the ViT-ARN model, which consists of two distinct models, each playing a specific role in anomaly detection and recognition in videos. 
The first model is an anomaly detection model based on one-class classification (OCC), aimed at predicting whether an anomaly is present in a video or not. This model relies on a VGG-type network to extract features from images, followed by a fully connected network to classify events as normal or anomalous. The second model is dedicated to recognizing the types of anomalies, using a Vision Transformer (ViT) to extract spatio-temporal features from videos. 
The images are divided into patches processed by a transformer encoder, and the process is refined by a Multi-Reservoir Echo State Network (MrESN). 
The final prediction is then made by a fully connected model.
Regarding hybrid models combining spatial and temporal analysis, Doshi and Yilmaz~\cite{doshi2020continual} proposed an approach combining YOLOv3 (non-retrained) for object detection and FlowNet2 for optical flow feature extraction.
These features are then processed by a KNN algorithm applied to surveillance camera images, as well as datasets such as CUHK Avenue, UCSD Pedestrian, and ShanghaiTech. However, it is worth noting that YOLOv3, although effective at its release, is now considered relatively outdated compared to newer versions like YOLOv7, which offer better performance and efficiency. 
Subsequently, they proposed MONAD (Multi-Objective Neural Anomaly Detector), an architecture composed of two main modules: 
\begin{itemize} 
\item The first module is a feature extraction module based on deep learning, using a generative adversarial network (GAN) to predict future video frames and compute the prediction error (MSE). This module also uses a lightweight object detector, YOLOv3, to extract localization information (center and area of the bounding box) and appearance information (class probabilities) of the objects detected in each frame. For each object, a feature vector is constructed by combining the prediction error, localization information, and class probabilities. 

\item The second module, dedicated to anomaly detection, uses a non-para-metric sequential algorithm to analyze feature vectors in real time. It compares new observations with normal training data using a k-nearest neighbors (KNN) approach~\cite{doshi2021online}. 
\end{itemize} 

However, using GANs to predict future frames is not ideal for real-time applications, as these models are often computationally expensive and introduce significant latency. 
More recently, Mostafa~\cite{ali2023real} introduced the AVAD (Autoencoder-based Video Anomaly Detection) method, which uses a convolutional autoencoder to detect abnormal frames and YOLOv5 (non-retrained) to identify objects responsible for anomalies. 
This method was applied to the same datasets used by Doshi and Yilmaz, including UCF Crime. However, the autoencoder-based approach has a significant limitation: the model must reconstruct each image in the input sequence, making it suboptimal for real-time applications due to the computational overhead associated with this process.

\subsection{Dataset}
Regarding existing datasets, Zhu et al.~\cite{zhu2021video} lists several datasets suitable for anomaly detection in videos. These include the Dashcam Accident Dataset (DAD), the Car Accident Dataset (CADP), A3D, DOTA Detection of Traffic Anomaly (DADA), UCSD, ShanghaiTech, and UCF Crime. 
These datasets can be categorized into three groups. On one hand, there are datasets that focus exclusively on a specific type of anomaly, such as DAD, CADP, A3D, and DADA, which are dedicated to traffic-related issues. 
On the other hand, datasets like UCSD and ShanghaiTech address low-impact anomalies related to safety, such as bicycles on sidewalks. 
Finally, the third category includes UCF Crime, which is the only dataset relevant to our study. It contains 1,900 raw videos divided into 13 types of anomalies: abuse, arrests, arson, assault, road accidents, burglary, explosions, fighting, armed robbery, shootings, shoplifting, theft, and vandalism, as well as videos without anomalies. 
However, despite its diversity, UCF Crime is insufficient for effectively training artificial intelligence models for video anomaly detection, as noted by Vrskova et al.~\cite{vrskova2022new}. 
In addition to being poorly cleaned and unbalanced, UCF Crime lacks a sufficient amount of data to meet the needs of anomaly detection models. Jacob~\cite{jacob2019anomaly} further points out that no dataset is currently rich enough to properly train a deep learning model for video anomaly detection. 
This view is reinforced by Samaila et al.~\cite{samaila2024}, who indicate that public datasets are limited in size and diversity, and most data is too unrealistic.

\section{Spatio-temporal Video Analysis Architecture}
\label{sec:proposition}
\begin{figure}[H]
    \centering
    \includegraphics[width=12cm]{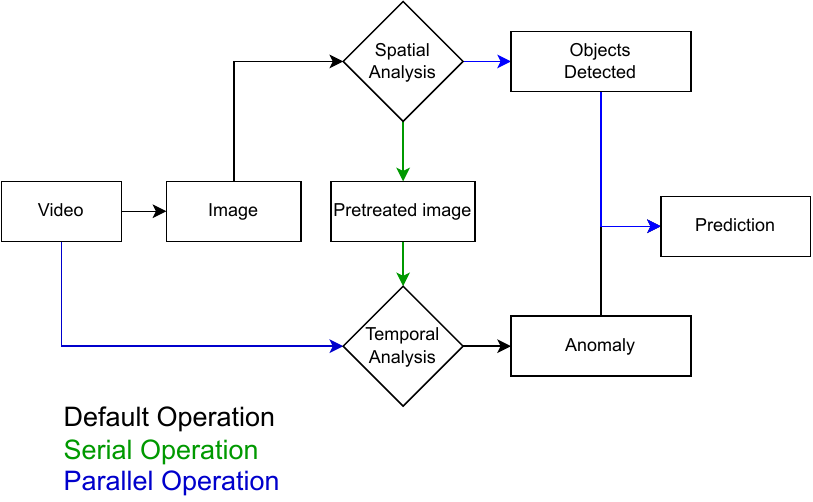}
    \caption{Spatio-temporal Video Analysis Architecture}
    \label{fig:Architecture}
\end{figure}

In this article, we propose an architecture composed of two complementary analyses: a spatial analysis and a temporal analysis. Figure~\ref{fig:Architecture} illustrates this dual approach and its overall structure. 

\begin{figure}[H]
    \centering
    \includegraphics[width=12cm]{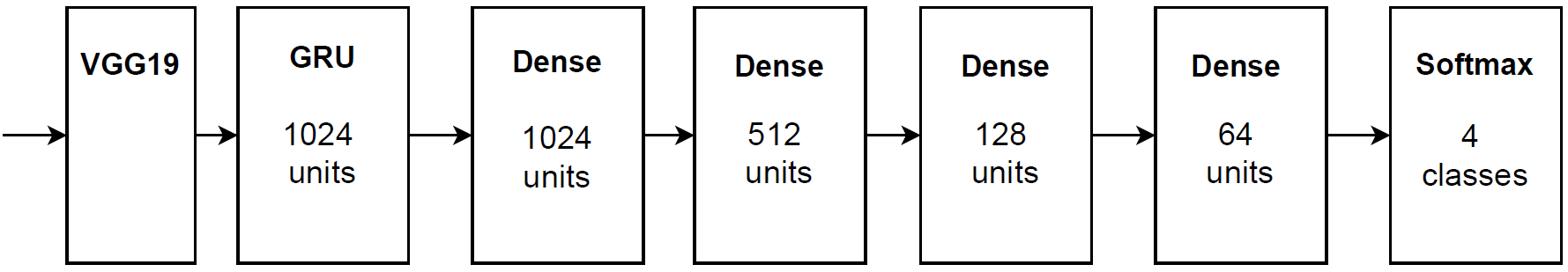}
    \caption{Structure of the Temporal Analysis Component VGG-GRU. \\The GRU layer has a dropout rate set to 50\%, as do all Dense layers, which also have L2 regularization fixed at 0.01.}
    \label{fig:CNN+GRU}
\end{figure}

\noindent On one hand, for spatial analysis, we have chosen to use YOLOv7 for its remarkable performance in terms of execution speed and detection accuracy.
The temporal analysis, on the other hand, relies on a neural network combining VGG19 (Visual Geometry Group) and GRU (Gated Recurrent Unit), also including an Multi-Layer Perceptron (MLP). This model is designed to process sequences of 15 images, each with a size of 112x112 pixels. VGG-GRU has already been used to achieve high performance in the detection of anomalies in videos~\cite{poirier2023enhancing}. Figure~\ref{fig:CNN+GRU} presents in detail the structure of this temporal analysis model. 

\noindent These two models have been trained on proprietary data sets\footnote{The data sets used are confidential and cannot be shared publicly.}. 
The data set for YOLOv7 consists of 10,000 images of firearms, 2,000 images of fires containing flames or smoke, people images, and images unrelated to these objects to ensure robust and adaptable detection. 
All images are in JPEG format. The VGG-GRU model, meanwhile, was trained on a corpus of MP4 videos representing three classes of anomalies: fights (978 videos), gunshots (311 videos), and fires (298 videos). These videos were carefully edited to retain only the segments that contained anomalies. Videos that did not contain anomalies were added to represent the``normal`` class. One of the main advantages of this architecture lies in its flexibility. The spatial analysis model can be positioned either in parallel or in series, depending on the specific needs of the application. 
In addition, by configuring the frame interval during the video preparation process, it is possible to handle video streams (continuous video) or downloaded videos (finished video). It is also possible to completely deactivate one of the two analysis modules or replace them with more recent versions without requiring the entire architecture to be retrained. For instance, a more advanced version of YOLO can be seamlessly integrated. 
Furthermore, the architecture can run on a CPU, although GPU execution is recommended for optimal performance.

\section{Experiments}
\label{sec:experiments}
In this section, we present various experiments that aim to compare the performance of our different architectures. We will begin by evaluating the architecture in which our components are arranged in parallel.

\subsection{Parallel Architecture}
\label{subsec:parallelArchi}
This architecture uses YOLOv7 to detect various key objects related to our anomalies, based on the assumption that the absence of these objects makes the occurrence of the anomaly highly unlikely. 
We focus on detecting people for the ``fight`` anomaly, the presence of flames and smoke for ``fire``, and the Intersection over Union (IoU) between a person and a firearm for the ``gunshot`` anomaly. Once predictions are made by our two analysis modules, they are combined using the following logical rule:

\begin{enumerate}
    \item If our model detects an anomaly, the predicted class will correspond to this anomaly.
    
    \item Otherwise, if a key object is detected, we will predict the anomaly associated with that object. However, in the specific case of a weapon, we will first check if the IoU between the weapon and a person is greater than 0 before predicting the "gunshot" anomaly.\footnote{Our model was designed to detect real weapons. In this context, no tests have been conducted on images containing toys, drawings, or other types of representations.}
\end{enumerate}

For the ``gunshot`` class, given that it involves two distinct objects and that a firearm only represents a risk when it is within reach of a person, we use the IoU between these two objects when combining the results of our detections. 
This approach allows for a more precise evaluation of the spatial context of the detected objects. Although the architectures presented in section \ref{sec:relatedWork} are diverse, they are not systematically comparable with one another. Most of them focus on unsupervised learning techniques or rely on models that do not necessarily meet the constraints we have set, such as real-time processing. Furthermore, while many architectures claim to be suitable for real-time anomaly detection, it is important to highlight that none of these studies provide specific timing metrics or performance benchmarks to support these claims.
For this reason, we chose to compare our model with C3D, a well-known architecture designed for a related task, namely action recognition, and for which pretrained weights are available. Our initial tests on our dataset demonstrate that our architecture, based on a combination of CNN + RNN + GRU, delivers slightly better performance than C3D, achieving an F1-score of 36.5\% compared to 35.7\%~\cite{poirier2023enhancing}. The results of our model are presented in Table~\ref{yoloCGRUmulti}.

\begin{table}[H]
 \begin{center}
 \caption{VGG-GRU + YOLO Performance}
\label{yoloCGRUmulti}
\vspace{0.3\baselineskip}
\begin{tabular}{cccc}
Accuracy & Precision & Recall & F1-Score \\ \hline
78.42\% & 85.60\% & 78.42\% & 81.16\% \\
\end{tabular}
\end{center} 

\begin{center}
\begin{tabular}{c|cccc}
\multicolumn{5}{c}{Confusion matrix (in percent)} \\ 
\diagbox{Truth}{Predicted} & Fight & Gunshot & Fire & Normal \\ \hline
Fight & \textbf{63.66\%} & 6.58\% & 1.93\% & 27.83\% \\
Gunshot & 9.94\% & \textbf{66.06\%} & 9.33\% & 14.67\% \\
Fire & 13.66\% & 15.73\% & \textbf{57.71\%} & 12.9\% \\ 
Normal & 7.43\% & 5.96\% & 3.98\% & \textbf{82.63\%} \\
\end{tabular}
\end{center}
\end{table}
Although the precision score presented in Table~\ref{yoloCGRUmulti} is 85.6\%, the recall (78.42\%) could be improved. These results demonstrate good anomaly detection but the confusion matrix reveal weaknesses for specific classes, such as the "fire" class, where confusion with other types of anomalies is high.

\subsection{Serial Architecture}
\label{subsec:serialArchi}
\begin{figure}[H]
    \centering
    \includegraphics[width=0.6\linewidth]{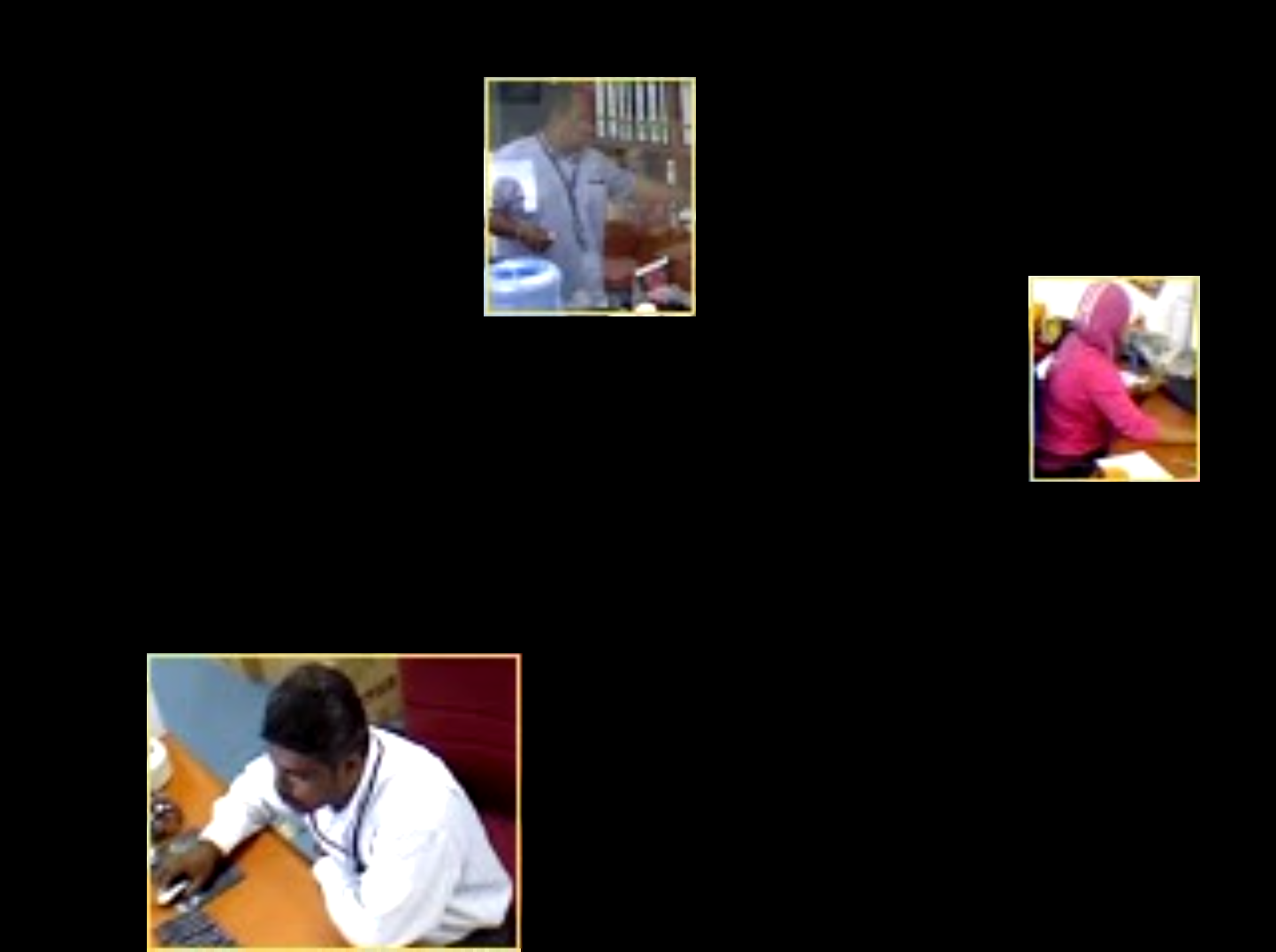}
    \caption{Example of Mask Generated with YOLO}
    \label{fig:masque}
\end{figure}

As part of our serial analysis, we leveraged YOLOV7 to enrich our input data by applying various preprocessing techniques. Our main approach consisted of removing the background from images, leaving only the key objects detected by YOLO. 
This method is illustrated in Figure~\ref{fig:masque}. 

\noindent The objective of this technique is to focus our model's attention on relevant elements, thus avoiding it from concentrating on parasitic background movements.
During this experimentation, we explored two options for cases where no key objects were detected by YOLO:
\begin{itemize}
    \item Retaining the original image without background: The results of this approach are presented in Table~\ref{tab:Masque_sans_fond_noir}.
    
    \item Introducing an entirely black image: This alternative is presented in Table~\ref{tab:Masque_fond_noir}.
\end{itemize}

\begin{table}[H]
 \begin{center}
 \caption{Performance for Mask without Black Background}
\label{tab:Masque_sans_fond_noir}
\vspace{0.3\baselineskip}
\begin{tabular}{cccc}
Accuracy & Precision & Recall & F1-Score \\ \hline
72.22\% & 82.37\% & 72.22\% & 75.68\% \\
\end{tabular}
\end{center} 

\begin{center}
\begin{tabular}{c|cccc}
\multicolumn{5}{c}{Confusion matrix (in percent) for mask without black background} \\ 
\diagbox{Truth}{Predicted} & Fight & Gunshot & Fire & Normal \\ \hline
Fight & \textbf{63.06\%} & 9.33\% & 1.96\% & 25.65\% \\
Gunshot & 25.99\% & \textbf{41.45\%} & 2.29\% & 30.27\% \\
Fire & 19.23\% & 15.98\% & \textbf{32.44\%} & 32.35\% \\ 
Normal & 15.33\% & 4.41\% & 2.07\% & \textbf{78.19\%} \\
\end{tabular}
\end{center}
\end{table}

\begin{table}[H]
 \begin{center}
 \caption{Performance for Mask with Black Background}
\label{tab:Masque_fond_noir}
\vspace{0.3\baselineskip}
\begin{tabular}{cccc}
Accuracy & Precision & Recall & F1-Score \\ \hline
75.58\% & 79.34\% & 75.58\% & 76.50\% \\
\end{tabular}
\end{center} 

\begin{center}
\begin{tabular}{c|cccc}
\multicolumn{5}{c}{Confusion matrix (in percent) for mask with black background} \\
\diagbox{Truth}{Predicted} & Fight & Gunshot & Fire & Normal \\ \hline
Fight & \textbf{55.52\%} & 4.43\% & 1.76\% & 38.29\% \\
Gunshot & 16.20\% & \textbf{41.44\%} & 1.53\% & 40.83\% \\
Fire & 14.09\% & 12.49\% & \textbf{20.59\%} & 52.83\% \\ 
Normal & 10.55\% & 2.98\% & 2.09\% & \textbf{84.38\%} \\
\end{tabular}
\end{center}
\end{table}

\begin{figure}[H]
    \centering
    \begin{minipage}[b]{0.45\textwidth}
        \centering
        \includegraphics[width=\textwidth]{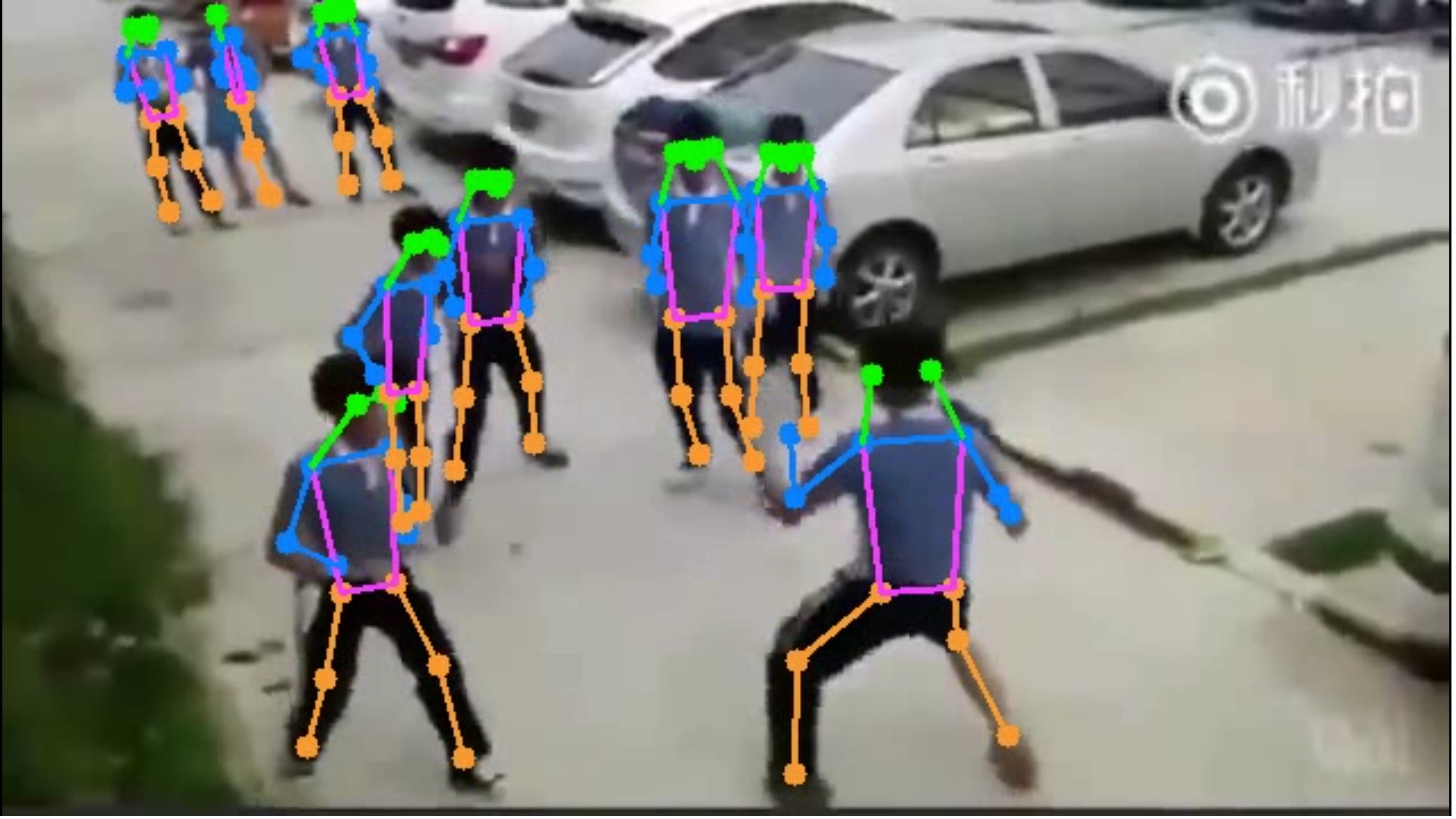}
        \caption{Pose Estimation by YOLOv7 with background}
        \label{fig:pose_fond}
    \end{minipage}
    \hfill
    \begin{minipage}[b]{0.45\textwidth}
        \centering
        \includegraphics[width=\textwidth]{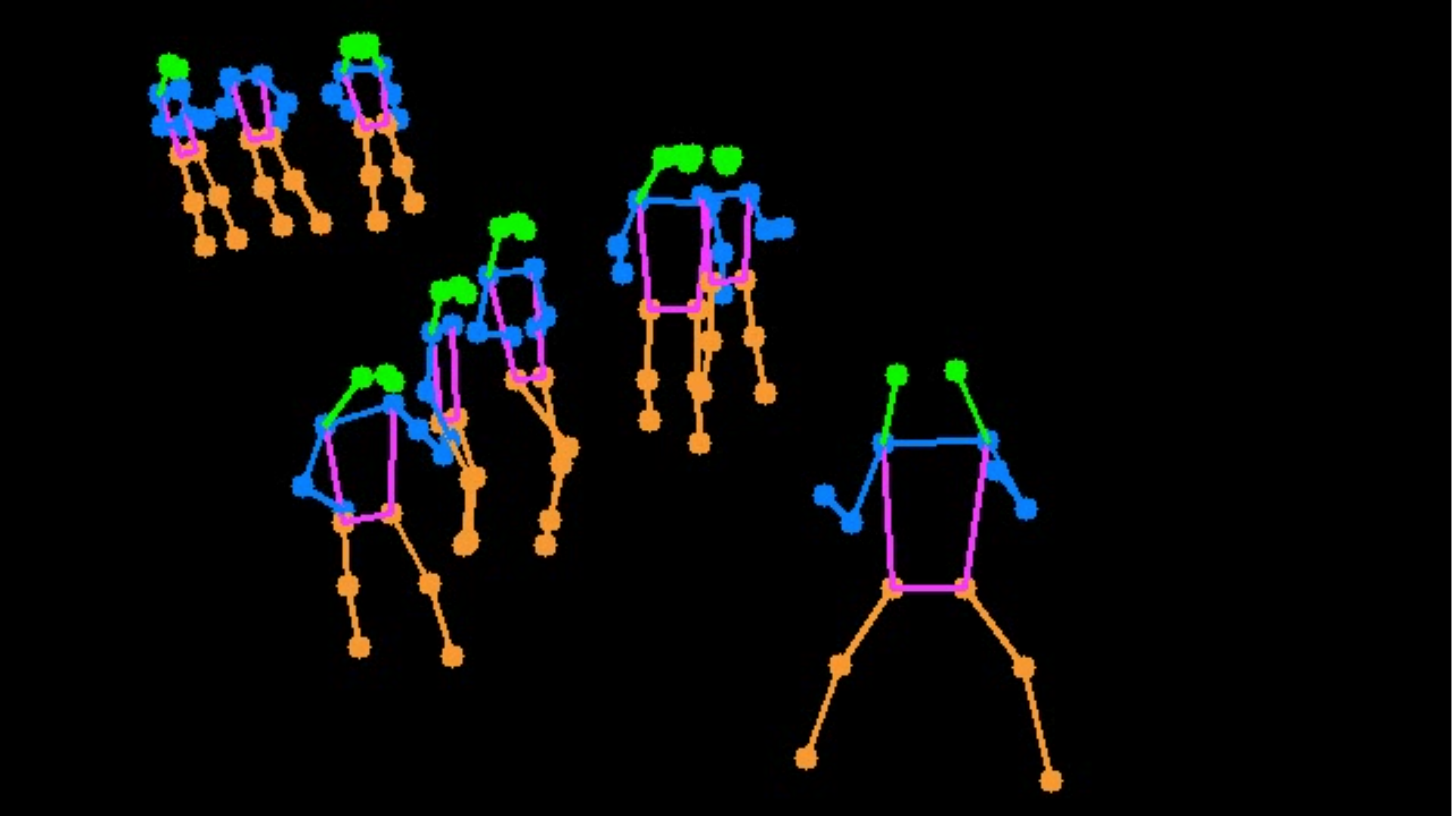}
        \caption{Pose Estimation by YOLOv7 without background}
        \label{fig:pose_sans_fond}
    \end{minipage}
\end{figure}

\noindent Comparing these two approaches, we noted that using a black image in the absence of detected objects improved the detection of the ``normal" class, increasing from 78\% to 84\%. 
However, this improvement came at the expense of precision for other classes, particularly ``fight" and "fire." 
For the ``fight" class, precision decreased from 63\% without a black image to 55\% (-8\%), while for the ``fire" class, it dropped from 32\% to 20\% (-12\%). 
Given the behavioral nature of certain anomalies, particularly those involving human interactions, we leveraged the flexibility of our architecture to integrate YOLOv7-pose as a replacement for standard YOLOv7. 
This adaptation allows us to trace the skeleton of each person present on the screen, thus offering a more refined analysis of movements and postures.
We experimented with two preprocessing approaches using YOLOv7-pose:

\begin{itemize}
    \item Preservation of the background with superimposition of detected skeletons: 
    Visual illustration (see Figure~\ref{fig:pose_fond}) 
    and Detailed results (see Table~\ref{pose_sans_fond_noir});
    
    \item Removal of the background, presenting only the skeletons on a black background: 
    Visual illustration (see Figure~\ref{fig:pose_sans_fond})  
    and Detailed results (see Table~\ref{pose_fond_noir}).
\end{itemize}

\begin{table}[H]
 \begin{center}
 \caption{YOLOv7-Pose + VGG-GRU without Background (3 Classes)}
\label{pose_fond_noir}
\vspace{0.35\baselineskip}
\begin{tabular}{cccc}
Accuracy & Precision & Recall & F1-Score \\ \hline
87.2\% & 90.9\% & 87.8\% & 89\% \\
\end{tabular}
\end{center} 

\begin{center}
\begin{tabular}{c|ccc}
\multicolumn{4}{c}{Confusion Matrix (in \%) for Pose Analysis without background} \\
\diagbox{Truth}{Predicted} & Fight & Gunshot & Normal \\ \hline
Fight & \textbf{64.5\%} & 1\% & 34.5\%  \\
Gunshot & 11.7\% & \textbf{70.7\%} & 17.6\% \\
Normal & 9.5\% & 0\% & \textbf{90.5\%}  \\
\end{tabular}
\end{center}
\end{table}

\begin{table}[H]
 \begin{center}
 \caption{YOLOv7-Pose + VGG-GRU with Background (4 Classes)}
\label{pose_sans_fond_noir}
\vspace{0.3\baselineskip}
\begin{tabular}{cccc}
Accuracy & Precision & Recall & F1-Score \\ \hline
87.3\% & 87.6\% & 87.3\% & 87.1\% \\
\end{tabular}
\end{center} 

\begin{center}
\begin{tabular}{c|cccc}
\multicolumn{5}{c}{Confusion Matrix (in \%) for Pose Analysis without Black Background} \\
\diagbox{Truth}{Predicted} & Fight & Gunshot & Fire & Normal \\ \hline
Fight & \textbf{60.5\%} & 2.4\% & 1.3\% & 35.8\% \\
Gunshot & 10\% & \textbf{55.6\%} & 14.8\% & 19.6\% \\
Fire & 15.5\% & 10.6\% & \textbf{48\%} & 25.9\% \\ 
Normal & 3.4\% & 0.6\% & 1\% & \textbf{95\%} \\
\end{tabular}
\end{center}
\end{table}

\noindent The results of these two experiments (see Table~\ref{pose_fond_noir} and Table~\ref{pose_sans_fond_noir}) revealed interesting trade-offs. 
The analysis of poses on a black background significantly improved the detection of anomalies related to human behaviors, such as fights (+4\%) and gunshots (+15\%), compared to the normal background. 
This improvement is explained by the increased focus of the model on people's movements and interactions, without distraction from the background. 
However, this approach also led to a notable decrease in the system's ability to detect fires. The removal of the background effectively eliminated crucial visual information for identifying flames and smoke, rendering these anomalies undetectable. 
This experiment highlights the importance of a judicious balance between focusing on human behaviors and preserving contextual information from the environment. It also underscores the need for an adaptive approach in anomaly detection, capable of adjusting based on the specific nature of the anomalies to be detected. Given the potential impact of the anomalies discussed in this article, we aimed to conclude our experiments by measuring various processing times. 
Each of these measurements was performed on a laptop equipped with 32 GB of RAM, an Intel Core i9 processor with 16 cores clocked at 2.3 GHz, and an Nvidia GeForce RTX2080 GPU with 8 GB of dedicated memory. 
As expected, the parallel architecture proved to be significantly faster than the sequential one (cf. Table~\ref{speed_parallele} and Table~\ref{speed_serie}). 

\begin{table}[H]
    \centering
    \caption{Execution time of YOLO and CGRU in parallel}
    \vspace{0.3\baselineskip}
    \label{speed_parallele}
    \begin{tabular}{c|c|c}
         Video duration & Average detections & Processing time  \\ \hline
         16s & 601ms & 15s \\
        44s & 533ms & 35s \\
        9s & 994ms & 12s \\
        35s & 1.1s & 57s \\
        23s & 1s06 & 35s \\
        1min 43 & 758ms & 116s (1min 56) \\
        50s & 826ms & 61s \\
        1min & 30 886ms & 83s (1min 23) \\
        2s & 847ms & 847ms \\
        9s & 870ms & 11s \\
        2s & 1s & 1s 
    \end{tabular}   
\end{table}

\begin{table}[H]
    \centering
     \caption{Execution time of YOLO and CGRU in serie}
    \label{speed_serie}
    \vspace{0.35\baselineskip}
    \begin{tabular}{c|c|c}
         Video duration & Average detections & Processing time  \\ \hline
         16s & 1s & 26s  \\
        44s & 1s & 71s (1min 11)  \\
        9s & 1.5s & 20s  \\
        35s & 1.5s & 81s (1min 21)  \\
        23s & 1.5s & 48s  \\
        1min 43 & 1.2s & 193s (3min 13)  \\
        50s & 1.3s & 102s (1min 42)  \\
        1min 05 & 1.4s & 134s (2min 14)  \\
        2s & 1.3s & 1.3s  \\
        9s & 1.3s & 17s  \\
        2s & 1.5s & 1.5s
    \end{tabular} 
\end{table}

\noindent This can be explained by the fact that, in the sequential setup, YOLOv7 must pre-process each video frame before being analyzed by our VGG-GRU model. However, the average prediction time for the sequential architecture remains reasonably fast, with an average of 1 to 1.5 seconds per prediction, allowing for a timely response when an anomaly is detected. 
It is worth noting that the execution speed of the parallel model can be further improved by adjusting the number of frames YOLOv7 needs to analyze. 
Due to the combination of both models, it is not necessary to analyze every frame in a sequence to make a prediction. Because of the redundant information between successive frames, the model can be configured to analyze a reduced number of frames per sequence, which enhances processing speed.

\section{Conclusion and Future works}
\label{sec:conclusion}
This paper proposes a hybrid architecture combining spatial and temporal analyses for real-time video anomaly detection. This approach leverages the strengths of YOLOv7 for object detection and the VGG-GRU model for temporal sequence analysis, offering flexibility in the arrangement of the modules according to specific needs. Our various experiments have allowed us to identify two optimal configurations for video anomaly detection, each addressing specific requirements:

\paragraph{For precise anomaly detection}
    The serial configuration, combining YO\-LOv7 with VGG-GRU, has proven particularly effective. This approach excels in identifying human behavioral anomalies. The integration of pose estimation preprocessing and background removal has significantly improved results, offering detailed analysis adapted to situations where accuracy is paramount.
    
\paragraph{For real-time analysis}
    The parallel architecture, combining YOLOv7 and our VGG-GRU, offers an optimal balance between reliability and speed. This configuration ensures prompt detection of anomalies while maintaining adequate accuracy, thus meeting the needs of applications requiring instant processing.

\vspace{0.5\baselineskip}
\noindent These results highlight the importance of an adaptive approach in video anomaly detection. Our modular architecture effectively responds to various scenarios, paving the way for diverse applications in the fields of security, surveillance, and event management. 
The flexibility of our system allows for prioritizing either precision or speed, depending on the specific requirements of each application.

\vspace{0.5\baselineskip}
\noindent Our study has highlighted several challenges and improvement perspectives for our anomaly detection system. Firstly, we observed that not all anomalies are necessarily linked to identifiable key objects, particularly in the case of natural disasters. Moreover, the presence of key objects does not always signify danger, as illustrated by the example of armed military personnel in airports. To overcome these limitations, we are considering directly transmitting the information collected by YOLO and VGG+GRU to our Multi-Layer Perceptron, allowing it to automatically learn the conditions for anomaly detection. We also noticed confusion between certain anomaly classes, suggesting the potential benefit of exploring a binary model to verify this hypothesis. 

\vspace{0.5\baselineskip}
\noindent Another promising approach would be to combine several different processing methods. 
For example, by associating a mask with the detected objects and representing people by their skeletons, which could improve the reliability of the sequential model, at the cost of reduced detection speed. Finally, to enhance the robustness and versatility of our system, it would be beneficial to enrich our dataset with new anomaly classes and examine the execution speed of our architecture compared to other existing models in the field.

\bibliographystyle{plain}
\bibliography{references}

\end{document}